\documentclass[runningheads]{llncs}
\usepackage[T1]{fontenc}
\usepackage{graphicx}
\usepackage{amsmath} 
\usepackage{xspace}
\usepackage{xcolor}
\usepackage[utf8]{inputenc}
\usepackage[linesnumbered,ruled,vlined]{algorithm2e}
\SetKwComment{Comment}{/* }{ */}
\usepackage{algpseudocode}
\usepackage{multirow}
\usepackage{enumitem}
\usepackage{pifont}
\usepackage{hyperref}
\usepackage{booktabs}

\newcommand{\exekgs}{\textit{Exe\-KGs}\xspace}
\newcommand{\exekg}{\textit{Exe\-KG}\xspace}
\newcommand{\mlseakg}{\textit{MLSea-KG}\xspace}
\newcommand{\mlsea}{\textit{MLSea}\xspace}

\newcommand{\newkg}{\textit{MetaExe-KG}\xspace}
\newcommand{\newframework}{\textit{KGmetaSP}\xspace}
\newcommand{\newbenchmark}{\textit{MetaExe-Bench}\xspace}
\newcommand{\kgmetaspdecomb}{\textit{KGmetaSP}\textsubscript{$DE_{comb}$}\xspace}
\newcommand{\kgmetaspdepip}{\textit{KGmetaSP}\textsubscript{$DE_{pip}$}\xspace}
\newcommand{\kgmetaspdevar}{\textit{KGmetaSP}\textsubscript{$DE_{var}$}\xspace}
\newcommand{\kgmetasppe}{\textit{KGmetaSP}\textsubscript{$PE$}\xspace}

\newcommand{\eg}[0]{\textit{e.g.},\xspace}
\newcommand{\ie}[0]{\textit{i.e.},\xspace}
\newcommand{\etal}[0]{et al.\xspace}
\newcommand{\dataentity}[0]{\texttt{DataEntity}\xspace}
\newcommand{\method}[0]{\texttt{Method}\xspace}
\newcommand{\task}[0]{\texttt{Task}\xspace}

\newcommand{\hyperparameter}[0]{\texttt{Hyper-parameter}\xspace}

\newcommand{\rdftvec}[0]{\textit{RDF2Vec}\xspace}
\newcommand{\mkga}[0]{\textit{MKGA}\xspace}
\newcommand{\dpse}[0]{\textit{DPSE}\xspace}
\newcommand{\ppe}[0]{\textit{PPE}\xspace}

\begin{document}

\title{Integrating Meta-Features with Knowledge Graph Embeddings for Meta-Learning}
\author{
Antonis Klironomos\inst{1,2}\thanks{These authors contributed equally to this work.}
\and
Ioannis Dasoulas\inst{3,4,5}\protect\footnotemark[1]
\and
Francesco Periti\inst{3,4,5}
\and
Mohamed H. Gad-Elrab\inst{1}
\and
Heiko Paulheim\inst{2}
\and
Anastasia Dimou\inst{3,4,5}
\and
Evgeny Kharlamov\inst{1,6}
}
\authorrunning{A. Klironomos and I. Dasoulas et al.}
\institute{
Bosch Center for Artificial Intelligence, Germany\\
\email{\{antonis.klironomos,mohamed.gad-elrab,evgeny.kharlamov\}@de.bosch.com}
\and
University of Mannheim, Germany\\
\email{heiko@informatik.uni-mannheim.de}
\and
KU Leuven, Belgium\\
\email{\{ioannis.dasoulas,francesco.periti,anastasia.dimou\}@kuleuven.be}
\and 
Leuven.AI - KU Leuven Institute for AI, Belgium \\
\and
Flanders Make, Belgium 
\and
Univesity of Oslo, Norway
}
\titlerunning{Integrating Meta-Features with Knowledge Graph Embeddings}

\maketitle

\begin{abstract}
The vast collection of experimental machine learning data available on the web presents a significant opportunity for meta-learning, where past experiments are leveraged to improve performance. 
Two crucial meta-learning tasks are \textit{pipeline performance estimation} (\ppe), which predicts pipeline performance on target datasets, and \textit{dataset performance-based similarity estimation} (\dpse), which identifies datasets with similar performance patterns. 
Existing approaches primarily rely on dataset meta-features
(\eg number of instances, class entropy, etc.) 
to represent datasets numerically and approximate these meta-learning tasks. 
However, these approaches often overlook the wealth of past experiment and pipeline metadata available. 
This limits their ability to capture dataset-pipeline interactions that reveal  performance similarity patterns. 
In this work, we propose \newframework, an approach utilizing knowledge graph embeddings which leverages existing experiment data to capture these interactions and improve both \ppe and \dpse.
We represent datasets and pipelines within a unified knowledge graph (KG) and derive embeddings that support pipeline-agnostic meta-models for \ppe and distance-based retrieval for \dpse.
To validate our approach, we construct a large-scale benchmark comprising 144,177 OpenML experiments, enabling a rich cross-dataset evaluation. 
\newframework enables accurate \ppe using a single pipeline-agnostic meta-model and improves \dpse over baselines.
The proposed \newframework, KG, and benchmark are released, establishing a new reference point for meta-learning and demonstrating how consolidating open experiment data into a unified KG advances the field.

\end{abstract}
\section{Introduction}

The rapid growth of data science has generated a vast collection of machine learning (ML) artifacts available on the web. 
Platforms such as OpenML~\cite{OpenML2013} host numerous records of experiments conducted across a wide range of datasets, pipelines and pipeline configurations.
A \textit{pipeline} is a data mining workflow composed of pre-processing, feature extraction, feature selection, estimation, and post-processing primitives, executed sequentially to prepare data and build predictive models~\cite{drori2021alpha3dm} and a \textit{pipeline configuration} is an instantiation of such pipeline in which all component hyper-parameters are specified.

The availability of ML experiment data on the web has sparked growing interest in meta-learning, a subfield of ML that leverages past experiments to improve performance~\cite{hospedales2022meta}. 
Two core meta-learning tasks are \textit{pipeline performance estimation (\ppe)}~\cite{carneiro2021using} and \textit{dataset performance-based similarity estimation (\dpse)}~\cite{leite2021exploiting}.
\ppe predicts pipeline configurations' performance for a target dataset, with predictive models, typically referred to as meta-models~\cite{carneiro2021using}.
\dpse measures how consistently datasets perform with shared pipeline configurations~\cite{feurer2014using}.
Solving these tasks helps identify promising pipelines early and narrow down the search space, reducing the need for costly training or optimization runs~\cite{feurer2014using,stolte2024methods}.

Existing methods for \ppe and \dpse typically rely on numeric dataset representations, \ie meta-features~\cite{castiello2005metadata,kalousis2001feature}, which summarize key dataset properties.
These include generic features (\eg number of instances), statistical features (\eg skewness), and landmarker features~\cite{bensusan2000discovering} (\eg classifier performance).
However, focusing exclusively on meta-features neglects a wealth of experimental metadata available on the web. 
This metadata can be modeled using semantic web technologies and represented within knowledge graphs (KGs), which capture the historical relationships between datasets and the ML pipelines evaluated on them.
These links enable \dpse even when datasets are evaluated with different pipeline configurations and provide historical dataset - pipeline usage patterns for \ppe. 
In this work, we examine whether encoding this information in a KG and learning numerical representations of its entities can benefit \ppe and \dpse.

\begin{figure}[t]
    \centering
    \includegraphics[width=.9\textwidth]{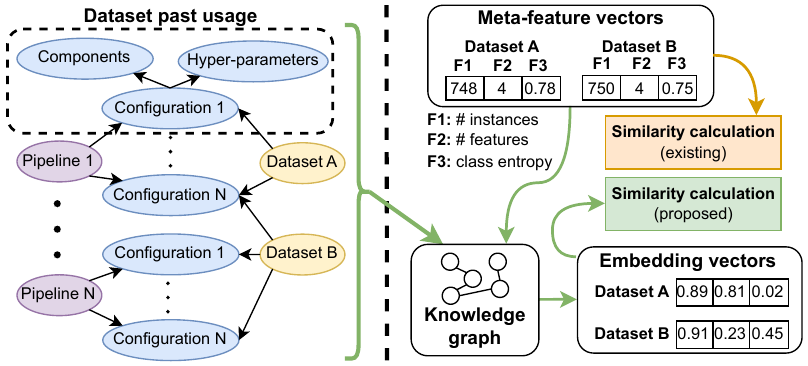}
    \vspace{-2ex}
    \caption{
    Comparison of the information leveraged by our proposed approach (green) versus existing approaches (orange) for representing datasets in meta-learning.
    For a pair of datasets, A and B, existing approaches represent datasets using only intrinsic meta-feature vectors (top-right). Our proposed approach also includes past pipeline configurations metadata (top left) as additional context into a KG, to represent datasets.
    }
    \label{fig:our-method-vs-existing}
\end{figure}

To this end, we propose \newframework, a novel approach that employs knowledge graph embeddings (KGEs) to model the latent structure relating datasets and their top-performing pipeline configurations (Fig.~\ref{fig:our-method-vs-existing}).
\newframework builds on the inherent ability of KGs to capture hidden patterns~\cite{nickel2016review}.
We construct \newkg, a unified KG that integrates semantic representations of datasets and their scikit-learn~\cite{pedregosa2011scikit} pipeline configurations extracted from OpenML~\cite{OpenML2013}.
For dataset representation, we adopt \mlseakg~\cite{dasoulas2024mlsea} to represent dataset metadata, including meta-features. 
For pipeline configuration representation, we adopt Executable KGs (\exekgs)~\cite{zheng2022executable}, which capture the data flow of ML pipeline configurations, including preprocessing steps, models, and hyper-parameters.
We employ KGE methods to learn a vector representation
for each dataset and pipeline configuration. 
We use pipeline embeddings to train pipeline-agnostic meta-models (\ie meta-models that are not tied to predictions for a specific pipeline configuration) for \ppe and use vector distance measures for \dpse. 

To evaluate our approach, we created a large-scale performance benchmark, namely \newbenchmark, comprising 144,177 scikit-learn pipeline evaluations.
To our knowledge, this is the first of its kind in terms of scale for existing OpenML datasets and pipelines, providing evaluation coverage for 2,616 end-to-end classification and regression pipeline configurations by executing them sparsely across 170 datasets.
First, we evaluated our proposed pipeline-agnostic meta-model for \ppe against commonly used specialized models across different prediction metrics for both unseen datasets and pipeline configurations.
Second, we compared our approach's \dpse by measuring distances between learned embeddings to baselines.  
Results demonstrate that \newframework enables accurate \ppe, allowing a single meta-model to jointly learn to predict the performance of multiple pipeline configurations, and leads to performance improvements in \dpse.
Our contributions are summarized as follows:\\(a) We present \newframework: a novel approach for assessing \ppe and \dpse using KGEs. 
It combines OpenML datasets with past pipeline configurations into a KG, from which embeddings are produced for use in downstream meta-learning tasks.
To our knowledge, this is the first investigation of KGEs for meta-learning tasks.
Our method enables accurate \ppe with a single meta-model and shows performance improvements in \dpse.\\
(b) We create a unified KG, \newkg, comprising 2,616 \exekgs derived from OpenML experiments on 170 datasets, covering classification and regression tasks. It integrates \exekgs~\cite{zheng2022executable} with \mlseakg~\cite{dasoulas2024mlsea} to include dataset meta-features, links between datasets and pipeline configurations, and detailed structural representations of pipeline configurations.\\
(c) We propose a large meta-learning benchmark, \newbenchmark, comprising 144,177 performance evaluations for 2,616 OpenML scikit-learn pipeline configurations across 170 datasets. 

\section{Related work}
\label{sec:related-work}

In this section we review relevant work for \ppe and \dpse. 
We also review the use of KGs in ML knowledge representation.

\medskip
\noindent \textbf{Pipeline Performance Estimation (\ppe).} 
Predicting a pipeline's performance for a target dataset is a core meta-learning task~\cite{lemke2013metalearning}.
Conventional approaches use the dataset’s meta-features as inputs for predictive meta-models, training on historical evaluations~\cite{carneiro2021using}. 
A common strategy involves training a distinct meta-model for each available pipeline configuration, predicting performance for unseen
datasets~\cite{bilalli2017on}.
While effective for warm-starting optimization~\cite{palumbo2023real}, these methods cannot generalize to pipeline configurations unseen during training since they do not encode pipelines numerically.
Pipeline embedding techniques such as DeepPipe~\cite{pineda2023deeppipe} mitigate this issue by modeling pipeline structure, but do not account for historical pipeline usage or dataset representations.
Our method jointly models datasets and pipelines, allowing \ppe both for unseen datasets and pipelines. 

\medskip
\noindent \textbf{Dataset Performance-based Similarity Estimation (\dpse).}
Identifying performance-similar datasets is a critical component of many meta-learning applications~\cite{stolte2024methods}.
Feurer \etal~\cite{feurer2015initializing} define dataset similarity based on how different pipeline configurations perform on them.
Prior work has proposed methods based on dataset meta-features~\cite{castiello2005metadata,kalousis2001feature}.
These can be simple (\ie descriptive measures such as number of instances), information-theoretic (\ie measures that quantify variable uncertainty and dependence)~\cite{10.5555/212782}, statistical (\ie measures that characterize the distribution of variables using moments and dispersion)~\cite{10.5555/212782}, and landmarkers (\ie measures derived by applying ML models on datasets and documenting their performance)~\cite{bensusan2000discovering}.
Other works learn meta-features by training neural networks on representations of entire datasets~\cite{jooma2021dataset,kim2018learning} based on a specific learning objective, generating task-specific meta-features. 
However, solely focusing on dataset representation overlooks valuable past experiment data. 
While landmarkers incorporate limited pipeline information, they require datasets to be evaluated on identical pipeline configurations to be comparable. 
Representing pipelines in the latent space allows \newframework to compare datasets even when evaluated with different pipeline configurations. 

\medskip
\noindent \textbf{Performance Benchmarks.}
Various large-scale benchmarking
efforts have been made to compare pipeline performance.
PMLB~\cite{olson2017pmlb} suite provides 165 classification
datasets sourced from repositories like UCI and KEEL and applies 13 models. 
AMLB~\cite{gijsbers2024amlb} offers a benchmark with 9 AutoML frameworks
on 71 classification and 33 regression datasets, mostly from OpenML.
Similarly, 
Alcobaça et al.~\cite{alcobaca2025exploring}
provides meta-knowledge from pipeline configuration variations
across 211 datasets retrieved from OpenML.
Still, these benchmarks rely on synthetic pipeline variations or evaluate only a limited subset of pipelines per dataset.
Our goal is to assess predictive performance using the pipeline configurations already available on the web, specifically, the scikit-learn pipelines recorded in OpenML, providing cross-dataset evaluations.

\medskip
\noindent \textbf{Ontologies and Graphs for ML Knowledge Representation.}
Several works use KGs to organize ML resources. HuggingKG~\cite{chen2025benchmarking} builds a KG of Hugging Face models and datasets.  MLTaskKG~\cite{liu2023taskoriented} constructs a KG of AI tasks and models with and their library implementations.
DCAT~\cite{dcat} provides a vocabulary for dataset metadata, while ML-Schema~\cite{publioMLSchemaExposingSemantics2018} and MLSO~\cite{dasoulas2024mlsea} extend this into the ML domain by describing datasets, tasks, and pipelines. \mlseakg~\cite{dasoulas2024mlsea} leverages MLSO to represent dataset metadata and associated pipelines, but does not model pipelines at the level of components. To standardize ML pipeline representation, \exekgs~\cite{zheng2022executable,klironomos2023exekglib}, capture the internal structure and data flow of ML pipelines, representing individual components and hyper-parameters.
Our \newkg is built upon the integration of \mlseakg and \exekgs to combine detailed ML dataset and pipeline representations.

\medskip
\noindent \textbf{Knowledge Graph Embeddings.}
KGEs learn low-dimensional vectors of nodes and relations in a KG for tasks, such as link prediction and node classification. KGE models are categorized as triple score-based, deep learning (\eg GNNs), and walk-based. Triple score-based methods use a scoring function~\cite{NIPS2013_1cecc7a7}, deep learning-based methods use neural networks~\cite{schlichtkrull2018modeling}, and walk-based methods generate node/relation sequences to learn vectors~\cite{ristoski2016rdf2vec}.
Score-based methods struggle with scalability and complex KGs~\cite{zheng2021pharmkg} while GNNs are computationally costly and assume homogeneous graphs~\cite{wu2020comprehensive}. 
Walk-based methods, such as \rdftvec~\cite{ristoski2016rdf2vec},
have been found to be more suitable for real-world sparse KGs~\cite{ristoski2019rdf2vec}.
\mkga~\cite{preisner2023universal} extends \rdftvec to include numeric literals by binning them.
Yet, KGEs' usage for downstream meta-learning tasks remains unexplored.

\section{Problem Description}
\label{sec:problem-description}


\medskip
\noindent \textbf{Context.}
Let $\mathcal{D}$ be a set of datasets, where each dataset $d \in \mathcal{D}$ is characterized by a set of meta-features $C_d$ (\eg number of instances). 
Let $\mathcal{P}$ be a set of ML pipeline configurations, where each pipeline configuration $p \in \mathcal{P}$ has been trained and evaluated on at least one dataset $d \in \mathcal{D}$ and has a set of characteristics $C_p$ (\eg preprocessing steps, predictive models, hyper-parameters).
Each pipeline configuration $p_j \in \mathcal{P}$ evaluated on a dataset $d_i \in \mathcal{D}$ produces a performance value $v_{i,j}$.

\medskip
\noindent \textbf{Goal.}
To constrain the pipeline search space in meta-learning, we consider \ppe and \dpse.
\ppe: for a target pair ($d_i, p_j$), the goal is to predict the performance value, denoted as $v_{i,j}$, that pipeline $p_j$ achieves on dataset $d_i$.
\dpse: for a target dataset $d_i \in \mathcal{D}$, the goal is to find a dataset $d_j \in \mathcal{D}$ ($j \neq i$) that exhibits the most similar performance profile across pipeline configurations $p \in \mathcal{P}$. 
For a dataset $d_i$, its performance profile is the vector of recorded performance values across the pipeline configurations evaluated on it. Formally, let $P_i = \{ p \in \mathcal{P} \mid p \text{ was evaluated on } d_i \}$; the performance profile is $V_i = [\, v_{i,1}, v_{i,2}, \ldots, v_{i,|P_i|} \,]$, where $v_{i,j}$ is the performance of pipeline $p_j$ on $d_i$.

\medskip
\noindent \textbf{Formalization.}
We formalize our two tasks. 
For \ppe, we define a meta-model $f$ that maps the characteristics of a dataset $d_i$ ($C_{d_i}$) and a pipeline $p_j$ ($C_{p_j}$) to a predicted performance value $\hat{v}_{i,j}$, such that it closely approximates the true value $v_{i,j}$:
$\hat{v}_{i,j} = f(C_{d_i}, C_{p_j})$.
For \dpse, we define a function $sim(d_i, d_j)$ that computes the similarity between two datasets $d_i, d_j$ based on $C_d$ and $C_p$. The retrieval problem is to find the dataset $d_j^*$ that maximises this function for a target dataset $d_i$:
$d_j^* = \underset{d_j \in \mathcal{D}, j \neq i}{\text{argmax}} \ sim(d_i, d_j)$.
\section{Knowledge Graph Embeddings for Meta-Learning}
\label{sec:methodology}

\begin{figure*}[t]
    \centering
    \includegraphics[width=0.9\textwidth]{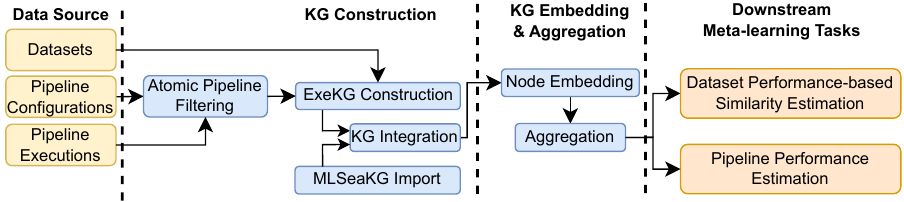}
    \vspace{-3ex}
    \caption{\newframework overview.
    }
    \label{fig:method-overview}
\end{figure*}

We propose \newframework, a novel meta-learning approach based on knoweledge graph embeddings (KGEs) (Fig.~\ref{fig:method-overview}).
KGEs offer a principled way to model similarity by encoding the relational topology among datasets, pipelines and their components in KGs.
By embedding this structure into a continuous space, KGEs place entities close together when they share similar relational contexts.

We start by constructing a unified KG, \newkg, that integrates both dataset-level and pipeline-level information. 
To this end, we combine \mlseakg~\cite{dasoulas2024mlsea}, which captures semantic and statistical properties of ML datasets, with ExeKGs we construct from OpenML experiments, which model the structure, components, and hyper-parameters of pipeline configurations.


We then embed \newkg to jointly represent datasets and pipelines in the latent space for downstream tasks, namely \ppe and \dpse.
Finally, we use pipeline embeddings to train pipeline-agnostic meta-models for \ppe and apply cosine similarity to dataset embeddings for \dpse.
Implementation details and computational cost analysis is provided in the Appendix.
We make our \newframework available, along with code to generate the \newkg~\footnote{\url{https://github.com/dtai-kg/KGmetaSP}}.

\subsection{\newkg: A Knowledge Graph of Datasets And Pipelines}

To support our approach, we require a KG that captures both rich dataset metadata and the structural details of ML pipeline configurations. 
MLSO, used in \mlseakg~\cite{dasoulas2024mlsea}, provides a detailed description of dataset characteristics, including metadata and meta-features.
However, MLSO lacks a representation of pipeline internals such as operators, components, and hyper-parameters. 
To fill this gap, we align \mlseakg with the ontology of \exekgs~\cite{zheng2022executable}, which offers a modular description of ML pipeline configurations. 
The resulting unified data model of our \newkg is shown in Fig.~\ref{fig:exekg-mlseakg-integration}.

\begin{figure}[t]
    \centering
    \includegraphics[width=1\textwidth]{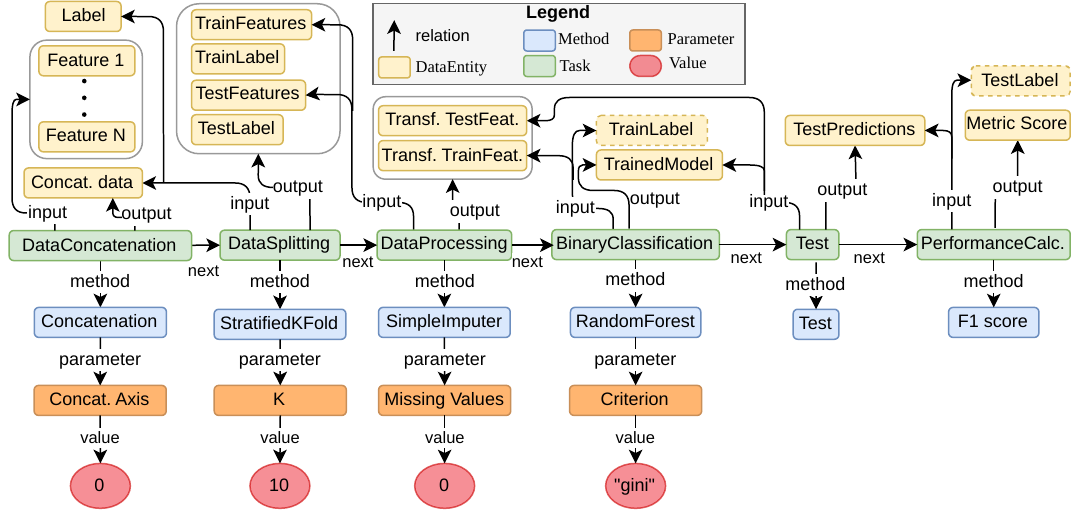}
    \vspace{-6ex}
    \caption{ML pipeline configuration example represented as an \exekg. 
    }
    \label{fig:exekg-example}
     \vspace{-2ex}
\end{figure}

Our data model is applied to datasets and pipeline configurations we fetch from OpenML. 
Specifically, we leverage the OpenML API\footnote{\url{https://www.openml.org/apis}} to fetch a set of OpenML datasets $\mathcal{D}$.
For each dataset $d_i \in \mathcal{D}$, we retrieve the top 10 pipeline configurations for each pipeline that has been trained and evaluated on $d_i$. 
Selecting the top configurations reduces noise from poorly performing runs, keeps the KG compact, and preserves the most informative historical performance patterns, improving embedding quality.
We focus on scikit-learn pipelines to maintain a consistent representation across pipeline components and hyper-parameters.
To ensure that the curated pipeline configurations can be executed end-to-end for their target dataset, we created a pipeline filtering module (Fig.~\ref{fig:method-overview}), to verify recorded OpenML evaluations and filter out pipeline configurations that are not atomic (\ie require additional preprocessing to be applied to the target dataset). 
We then apply MLSO and \exekgs' ontology to model datasets and pipeline configurations, including all their discrete steps and hyper-parameters as seen in Fig.~\ref{fig:exekg-example}.
This approach results in linking each dataset with its top-performing pipeline configurations recorded in OpenML.

For each OpenML dataset, its features and target variable are mapped to the corresponding \exekg semantic representations (Fig.~\ref{fig:exekg-example}).
For each associated pipeline configuration, we parse each step of their pipeline and the corresponding hyper-parameters.
We create an \exekg \dataentity node for each dataset variable, an \exekg \task node for each pipeline step (\eg data splitting, data transformation, model training and testing) and an \exekg \method node that implements the \exekg \task. 
This results in an independent \exekg graph for each pipeline configuration we mine. 

The pipeline configuration graphs are further enriched by aligning them with \mlseakg, which provides granular metadata for OpenML datasets, such as dataset descriptions, keywords and meta-features. 
This integration creates a unified knowledge representation that combines execution-level details with rich dataset characterization, resulting in \newkg.
Currently, \newkg includes 170 OpenML datasets and 2,616 OpenML (classification and regression) pipeline configurations, with an average of 15 steps per pipeline. In total, \newkg contains approximately 4.5 million triples. 

\subsection{\newframework: Knowledge Graph Embeddings for Meta-Learning}

After constructing \newkg, we embed its nodes into a continuous vector space using \rdftvec~\cite{ristoski2016rdf2vec} with \mkga~\cite{preisner2023universal} for binning numeric attributes.
We produce a single vector representation for each pipeline configuration and each dataset by aggregating the embeddings of their related nodes, exploring three aggregation strategies for datasets.
The learned embeddings are used for \ppe and \dpse as explained in the remainder of the section.

\begin{figure*}[t]
    \centering

    \begin{minipage}[t]{0.49\textwidth}
        \centering
        \includegraphics[width=\textwidth]{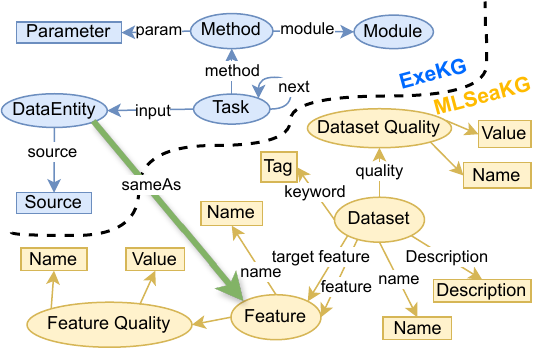}
        \vspace{-6ex}
        \caption{High-level view of \newkg.}
        \label{fig:exekg-mlseakg-integration}
    \end{minipage}
    \hfill
    \begin{minipage}[t]{0.49\textwidth}
        \centering
        \includegraphics[width=\textwidth]{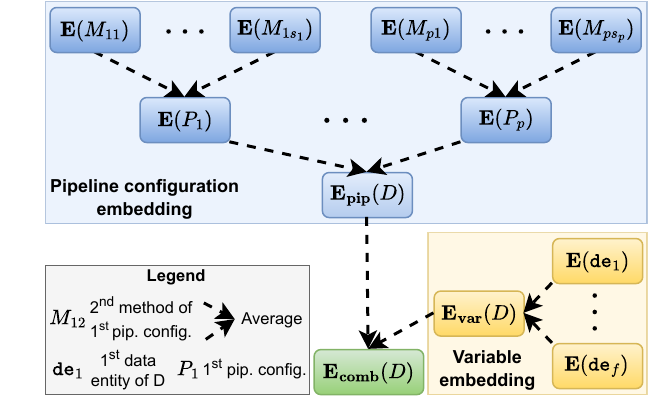}
        \vspace{-6ex}
        \caption{Embedding aggregation process.}
        \label{fig:kge-aggregation}
    \end{minipage}

\end{figure*}

\medskip
\noindent 
\textbf{Embedding Aggregation.} 
For each dataset, we aggregate node embeddings $E$ to create two representations (1, 2) based on different node types and a third, composite representation (3) (Fig.~\ref{fig:kge-aggregation}):
(1) $\mathbf{DE_{var}}$, by averaging \dataentity node embeddings, which represent dataset variables; (2) $\mathbf{DE_{pip}}$, by averaging \method node embeddings comprising the associated pipeline configurations; and (3) $\mathbf{DE_{comb}}$, by averaging $\mathbf{DE_{var}}$ and $\mathbf{DE_{pip}}$.
These two node types were chosen because they capture the main factors known to drive performance behaviour: the dataset's intrinsic properties, reflected at the variable level \cite{brazdil2008meta,kalousis2001feature}, and its empirical interaction with modeling components, reflected at the method level \cite{feurer2015initializing,leite2021exploiting}. In contrast, \hyperparameter nodes capture execution-specific settings, and \task nodes describe only data-flow stages. They do not convey the actual operation applied to the data, which is defined at the method level together with linked hyper-parameters.
$\mathbf{DE_{comb}}$ combines the two complementary perspectives, a common method for computing meta-embeddings~\cite{coates-bollegala-2018-frustratingly}. We define:
\begin{align*}
\mathbf{DE_{var}} (D_i) &= \frac{1}{f}\sum_{j=1}^{f} \mathbf{E}(\texttt{de}_j), &  \mathbf{PE} (P_l) &= \frac{1}{s_l} \sum_{k=1}^{s_l} \mathbf{E}(M_{lk}),\\
\mathbf{DE_{pip}} (D_i) &= \frac{1}{p} \sum_{l=1}^{p} \mathbf{PE} (P_l), & \mathbf{DE_{comb}} (D_i) &= \frac{1}{2} (\mathbf{DE_{var}}(D_i) + \mathbf{DE_{pip}}(D_i))
\end{align*}
where $\mathbf{E}(\texttt{de}_j)$ is the embedding of the $j$-th \dataentity node for dataset $D_i$; $f$ is the number of \dataentity nodes for $D_i$; $\mathbf{E}(M_{lk})$ is the embedding of the $k$-th \method node, in the $l$-th pipeline configuration associated with $D_i$; $s_l$ is the number of \method nodes in the $l$-th pipeline configuration associated with $D_i$, and; $p$ is the number of pipeline configurations associated with $D_i$.



\medskip
\noindent \textbf{\ppe and \dpse leveraging KGEs.} 
For \ppe, we use pipeline configuration embeddings ($\mathbf{PE} (P_l)$) to train pipeline-agnostic meta-models, which are not tied to a specific pipeline configuration, by combining them with dataset metafeatures (Fig.~\ref{fig:meta-models}).
For \dpse, we approximate dataset similarity of two datasets $D_1$ and $D_2$ through the cosine similarity of their embeddings, $\mathbf{DE}_{*}^{(1)}$ and $\mathbf{DE}_{*}^{(2)}$, where $*$ indicates the aggregation strategy ($\mathbf{DE_{var}}$, $\mathbf{DE_{pip}}$, or $\mathbf{DE_{comb}}$).
\section{Experiments}
\label{sec:experiments}

We evaluate our approach for \ppe and \dpse.
To this end, we construct \newbenchmark a ground-truth benchmark, obtained by executing \newkg pipeline configurations across \newkg datasets, thus
providing real performance values for training meta-models and comparing similarity estimates.

\subsection{Experimental Setup}

\medskip
\noindent \textbf{Data Source and \newbenchmark.}
Table~\ref{tab:kg-stats} summarizes the statistics of our data source with and without integrating \mlseakg, allowing us to assess the impact of semantic enrichment. 
To avoid leakage, we remove performance values, curated from OpenML, from \newkg before generating embeddings. 
Including them might lead the embedding model to encode performance information directly to the embeddings, instead of capturing the structural patterns. 

To evaluate our approach, we created \newbenchmark: a large-scale simulation benchmark obtained by sparsely training and evaluating the 2,616 \newkg pipeline configurations across the 170 \newkg datasets.
To achieve this, we mapped OpenML pipeline components and hyper-parameter metadata to scikit-learn \texttt{Pipeline}
objects. 
Due to inconsistencies between different scikit-learn versions of OpenML pipelines, we mapped deprecated parameters and outdated components of older configurations to the modern scikit-learn v1.5.1. 
The sparse simulation produced 144,177 distinct experiments of existing OpenML datasets and pipeline configurations. 

For each dataset $d_i \in D$, we train and evaluate a set of pipeline configurations $P$ compatible with its ML goal (\eg classification).
If $p_j$ cannot be used to fit $d_i$, the experiment is deemed as invalid and $V_{i,j}$ is assigned the worst possible value, depending on the performance metric. 
In addition, we measure fit time and penalize pipeline configurations that exhibit long execution times compared to their original execution, by deeming them invalid. 
For this version of the benchmark, we deem pipeline configuration that require at least 10 times the execution time of the original execution as invalid.

\begin{figure}[t]
    \centering
    \includegraphics[width=1\textwidth]{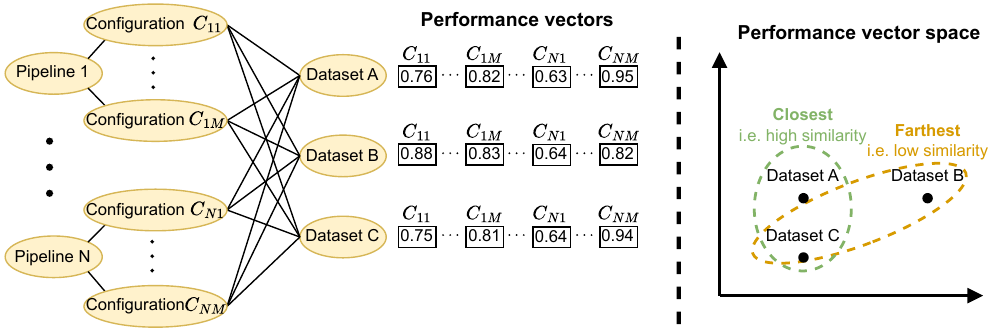}
    \vspace{-6ex}
    \caption{Overview of \newbenchmark constructed by sparsely training and evaluating \newkg datasets with \newkg pipeline configurations. Resulting performance values are used to assess the ground truth performance-based similarity between datasets and train \ppe meta-models.
    }
    \label{fig:perf-bench}
\end{figure}

For \ppe, we use the performance metric obtained when a dataset is evaluated with a pipeline configuration as the ground-truth target for prediction.
For \dpse, we derive the ground-truth similarity between two datasets from their performance profiles across the pipeline configurations on which both were trained and evaluated (Fig.~\ref{fig:perf-bench}). 
We compute the cosine similarity of their performance vectors~\cite{leite2021exploiting}, and use it as the performance-based similarity reference.

\newbenchmark is open-sourced and available for the ML community: {\url{https://github.com/dtai-kg/KGmetaSP}}.

\begin{table}[t]
    \caption{\newkg statistics.}
    \vspace{-2ex}
    \centering
    \setlength{\tabcolsep}{3pt}
    \begin{tabular}{lrrrrr}
        \toprule
        & \textbf{Entities} & \textbf{Properties} & \textbf{Triples} & \textbf{Attributes} & \textbf{Literals} \\
        \midrule
        \textbf{\exekgs} & 1,502,140 & 293 & 4,492,036 & 149 & 371,534 \\
        \textbf{\exekgs+\mlsea} & 1,514,836 & 317 & 4,522,216 & 155 & 385,246 \\
        \bottomrule
    \end{tabular}
    \label{tab:kg-stats}
\end{table}

\medskip
\noindent \textbf{Pipeline Performance Estimation (\ppe) Evaluation.}
We evaluate \newframework's ability on \ppe, using its embeddings as input to a single, pipeline-agnostic meta-model that predicts performance for any dataset - pipeline configuration pair.
While our \newframework supports any predictive meta-model, for our experiments we adopt random forest predictors, a standard choice in meta-learning~\cite{carneiro2021using}.  
We focus on pipeline configurations solving classification tasks and use the accuracy and precision values from \newbenchmark as the targets to be predicted (meta-targets). 
We assess the performance of \newframework in predicting the raw meta-targets (meta-regression) and classifying them in bins (meta-classification).
For meta-classification, meta-targets  are binned into three proportional intervals (low, medium, and high performance) using a quantile-based approach, ensuring an even distribution of samples across the bins.

\begin{figure}[t]
    \centering
    \includegraphics[width=1\textwidth]{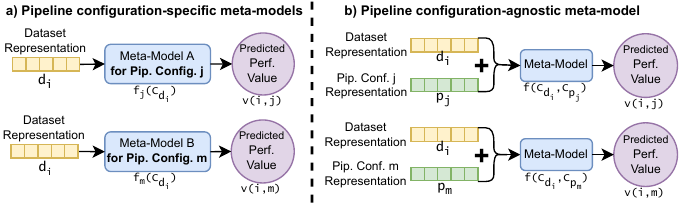}
    \vspace{-6ex}
    \caption{High-level depiction of \ppe meta-models. a) Common approach: Pipeline configuration-specific meta-models predict the performance of a pipeline-configuration for a dataset given a dataset's representation. b) Our approach: A single common pipeline configuration-agnostic meta-model predicts the performance of a pipeline configuration for a dataset given the dataset's and pipeline configuration's representations.}
    \label{fig:meta-models}
\end{figure}

We evaluate \ppe for unseen datasets and unseen pipeline configurations.
For \textbf{unseen datasets}, the goal is to predict the performance of pipeline configurations on datasets not observed during training.
We consider pipeline configurations that appear in at least 50 training samples (\ie dataset - pipeline pairs), totaling 123,920 samples from \newbenchmark across 1,028 pipeline configurations.
We compare our pipeline-agnostic meta-model which takes as input the concatenation of dataset meta-features and \newframework's pipeline configuration embeddings against 1,028 pipeline configuration-specific meta-models trained on dataset meta-features (Fig.~\ref{fig:meta-models}).
We evaluate different types of meta-features (MF), to test our meta-model's robustness:

\begin{itemize}[topsep=3pt,parsep=0pt,partopsep=0pt,itemsep=2pt,leftmargin=*]
\item \textbf{MF All:} A complete set combining simple, statistical, information-theoretic, and landmarker meta-features.
\item \textbf{MF Statistical:} Meta-features derived from the dataset's statistical properties, including measures of central tendency or dispersion~\cite{10.5555/212782}.
\item \textbf{MF Information Theory:} Meta-features derived from information theory concepts like entropy or mutual information~\cite{10.5555/212782}.
\item \textbf{MF Landmarkers:} Meta-features based on the performance of simple ML models (landmarkers) like Decision Tree or Naive Bayes on the dataset~\cite{bensusan2000discovering}.
\end{itemize}

For \textbf{unseen pipeline configurations}, the goal is to predict the performance of pipeline configurations not encountered during training, a scenario where configuration-specific meta-models cannot generalize, since they are trained to make predictions for a single pipeline configuration (Fig.~\ref{fig:meta-models}). 
We compare against two baselines that are both generalizable and pipeline-agnostic.
``Average performance'' is a naive constant predictor that predicts the average performance across all training samples.
``Closest embedding'' is a stronger retrieval-based heuristic that
uses the nearest observed pipeline in the embedding space.
These baselines bracket the task’s difficulty, ranging from ignoring all pipeline information to relying only on local similarity. Together, they provide a meaningful reference point for assessing \newframework's ability to generalize.

\medskip
\noindent \textbf{Dataset Performance-based Similarity Estimation (\dpse) Evaluation.}
We evaluate \newframework's effectiveness in \dpse.
First, we compute the ground-truth similarity for each dataset pair using \newbenchmark, based on the cosine similarity of their performance profiles.
We then compare the retrieval and ranking ability of our method against this ground truth using the $Hit@K$ and $NDCG@K$ metrics.
For $Hit@K$, a hit is recorded when a dataset whose ground-truth similarity with the query dataset exceeds $0.9$ appears among the top-$K$ retrieved datasets.
For comparison, we employ the below non-KGE baseline methods for \dpse from literature.

\begin{itemize}
[topsep=3pt,parsep=0pt,partopsep=0pt,itemsep=2pt,leftmargin=*]
\item \textbf{Graph Edit Distance (GED).} We use GED~\cite{redyuk2024assisted} to measure similarity between two datasets 
($d_1$, $d_2$) via the structural similarity of their pipeline configurations ($P_1$, $P_2$): $
\text{Sim}_{GED}(d_1, d_2) = -\frac{1}{|P_1| \cdot |P_2|} \sum_{p_i \in P_1} \sum_{p_j \in P_2} GED(p_i, p_j)$.

\item \textbf{Meta-feature P-Norm Distance Ranking (MF).} We adopt past works based on meta-feature distance~\cite{feurer2015initializing}. 
We employ the same set of meta-features as in \mlseakg, and test with and without meta-feature normalization. 
The similarity between datasets, $d_1$ and $d_2$, is calculated from the p-norm of their meta-feature vectors, $m_1$ and $m_2$: $\text{Sim}_{MF}(d_1, d_2) = - ||m_1 - m_2||_p$.

\item \textbf{Single-field Document Ranking: Dataset Meta-features (SiFi-Dat.).} The effectiveness of representing and ranking tables as documents has been shown in past works~\cite{cafarella2008webtables,zhang2018ad}. 
We investigate this in the context of \dpse, representing OpenML dataset metadata, such as title, description, and variable names as a single document and leveraging pre-trained bi-encoder language model embeddings with cosine similarity for ranking~\cite{reimers2019sentence}.

\item \textbf{Single-field Document Ranking: Pipeline Description (SiFi-Pip.).} We represent OpenML descriptions of top-performing pipeline configurations per dataset as a single document. 
Documents are ranked with pre-trained bi-encoder language model embeddings leveraging cosine similarity~\cite{reimers2019sentence}.

\item \textbf{Multi-field Document Ranking (MuFi-Dat.).} We represent each dataset as a fielded document~\cite{pimplikar2012answering,zhang2018ad} using metadata fields: title, description, variable names, and tags. 
We use the mixture of language models~\cite{ogilvie2003combining} approach, empirically setting field weights and rank embeddings using cosine similarity. 

\end{itemize}

\subsection{Evaluation Results}


\medskip
\noindent \textbf{Evaluation Results for \ppe.}
For the \textbf{unseen datasets} scenario (Table~\ref{tab:run-perf-pred-combined-unseen-datasets-compact}), in the meta-regression task, while configuration-specific meta-models show generally lower MSE and higher R2, our meta-model maintains comparable performance.
Notably, for the more practical task of meta-classification, our meta-model shows a distinct advantage, consistently outperforming the average of 1,028 specialized models across all dataset meta-feature types (MF) (\eg up to 0.7537 accuracy for ``MF All'').
This suggests that while our \newframework is less precise for point-value prediction, its strength lies in correctly ranking and identifying high-performing pipeline configurations, which is often the primary goal in meta-learning.
This is achieved with a single meta-model, offering significant efficiency gains over training and maintaining hundreds of specialized meta-models.
The pipeline embeddings' contribution is maintained for all tested dataset representations, showcasing the robustness of our approach.
In \ppe for \textbf{unseen pipeline configurations}, our meta-model significantly outperforms both ``Average performance'' and ``Closest embedding'' baselines (Table~\ref{tab:run-perf-pred-combined-unseen-runs-compact}).
These results highlight the strong generalization capabilities of our approach, enabling reliable \ppe even for pipeline structures not seen during meta-model training.

\begin{table}[t]
\centering
\caption{\ppe meta-regression and meta-classification for unseen datasets. We compare a single, configuration-agnostic model trained on meta-features and \newframework embeddings against an average evaluation of 1,028 configuration-specific specialized models trained on meta-features. Meta-regression targets are raw performance values.
Meta-classification targets are binned into 3 classes (low, medium, high). The highest score per metric is underlined; the best score per dataset embedding group is bold.
}
\label{tab:run-perf-pred-combined-unseen-datasets-compact}
\setlength{\tabcolsep}{2pt} 
\begin{tabular}{@{}llrrrrr@{}}
\toprule
\textbf{Dataset} & \textbf{Pipeline} & \textbf{Meta-} & \multicolumn{2}{c}{\textbf{Meta-Regr.}} & \multicolumn{2}{c}{\textbf{Meta-Classif.}} \\
\cmidrule(lr){4-5} \cmidrule(lr){6-7}
\textbf{Embedding} & \textbf{Strategy} & \textbf{Models} & \textbf{MSE} & \textbf{R2} & \textbf{Accuracy} & \textbf{F1} \\
\midrule
\multicolumn{6}{l}{\textbf{Target: Accuracy}} \\
\multirow{2}{*}{MF All} & Conf.-specific & 1,028 & \textbf{\underline{0.0081}} & \textbf{\underline{0.6748}} & 0.7363 & 0.7358 \\
& \textbf{\kgmetasppe} & 1 & 0.0101 & 0.6181 & \textbf{\underline{0.7413}} & \textbf{\underline{0.7427}} \\
\cmidrule(l){2-7}
\multirow{2}{*}{MF Info. Theory~\cite{10.5555/212782}} & Conf.-specific & 1,028 & \textbf{0.0249} & \textbf{-0.0737} & 0.4428 & 0.4423 \\
& \textbf{\kgmetasppe} & 1 & 0.0401 & -0.5135 & \textbf{0.4773} & \textbf{0.4789} \\
\cmidrule(l){2-7}
\multirow{2}{*}{MF Landmarkers~\cite{bensusan2000discovering}} & Conf.-specific & 1,028 & \textbf{0.0092} & \textbf{0.6428} & 0.7152 & 0.7139 \\
& \textbf{\kgmetasppe} & 1 & 0.0104 & 0.6094 & \textbf{0.7169} & \textbf{0.7172} \\
\cmidrule(l){2-7}
\multirow{2}{*}{MF Statistical~\cite{10.5555/212782}} & Conf.-specific & 1,028 & \textbf{0.0218} & \textbf{0.0892} & 0.5499 & 0.5484 \\
& \textbf{\kgmetasppe} & 1 & 0.0259 & 0.0212 & \textbf{0.5827} & \textbf{0.5830} \\
\midrule
\multicolumn{6}{l}{\textbf{Target: Precision}} \\
\multirow{2}{*}{MF All} & Conf.-specific & 1,028 & \textbf{\underline{0.0133}} & \textbf{\underline{0.6311}} & 0.7347 & 0.7400 \\
& \textbf{\kgmetasppe} & 1 & 0.0164 & 0.5449 & \textbf{\underline{0.7537}} & \textbf{\underline{0.7562}} \\
\cmidrule(l){2-7}
\multirow{2}{*}{MF Info. Theory~\cite{10.5555/212782}} & Conf.-specific & 1,028 & \textbf{0.0330} & \textbf{-0.0933} & 0.4324 & 0.4350 \\
& \textbf{\kgmetasppe} & 1 & 0.0506 & -0.4059 & \textbf{0.4961} & \textbf{0.5007} \\
\cmidrule(l){2-7}
\multirow{2}{*}{MF Landmarkers~\cite{bensusan2000discovering}} & Conf.-specific & 1,028 & \textbf{0.0148} & \textbf{0.6069} & 0.6987 & 0.7006 \\
& \textbf{\kgmetasppe} & 1 & 0.0184 & 0.4889 & \textbf{0.7247} & \textbf{0.7269} \\
\cmidrule(l){2-7}
\multirow{2}{*}{MF Statistical~\cite{10.5555/212782}} & Conf.-specific & 1,028 & \textbf{0.0274} & \textbf{0.1221} & 0.5490 & 0.5487 \\
& \textbf{\kgmetasppe} & 1 & 0.0346 & 0.0381 & \textbf{0.5897} & \textbf{0.5896} \\
\bottomrule
\end{tabular}
\end{table}

\begin{table}[t]
\centering
\caption{\ppe meta-regression and meta-classification for unseen pipeline configurations. Meta-Regression targets are the raw performance values.
Meta-classification targets are binned into 3 classes (low, medium, high). The best scores are in bold.}
\vspace{-2ex}
\label{tab:run-perf-pred-combined-unseen-runs-compact}
\setlength{\tabcolsep}{3pt} 
\begin{tabular}{lrrrr}
\toprule
\textbf{Method} & \multicolumn{2}{c}{\textbf{Meta-Regression}} & \multicolumn{2}{c}{\textbf{Meta-Classification}} \\
\cmidrule(lr){2-3} \cmidrule(lr){4-5}
& \textbf{MSE} & \textbf{R2} & \textbf{Accuracy} & \textbf{F1} \\
\midrule
\multicolumn{5}{l}{\textbf{Target: Accuracy}} \\
Average performance & 0.0267 & -0.0005 & 0.3303 & 0.1640 \\
Closest embedding & 0.0127 & 0.5241 & 0.7748 & 0.7747 \\
\textbf{MF All + \kgmetasppe} & \textbf{0.0070} & \textbf{0.7361} & \textbf{0.8250} & \textbf{0.8244} \\
\midrule
\multicolumn{5}{l}{\textbf{Target: Precision}} \\
Average performance & 0.0352 & -0.0012 & 0.3323 & 0.1658 \\
Closest embedding & 0.0225 & 0.3607 & 0.7773 & 0.7771 \\
\textbf{MF All + \kgmetasppe} & \textbf{0.0131} & \textbf{0.6264} & \textbf{0.8251} & \textbf{0.8247} \\
\bottomrule
\end{tabular}
\end{table}

\medskip
\noindent \textbf{Evaluation Results for \dpse.}
Tables~\ref{tab:results-hit} and~\ref{tab:results-ndcg} show the aggregation strategies of \newframework yield strong retrieval performance.
The combined strategy (\kgmetaspdecomb) achieves the best Hit@1 (tied with SF-Dataset), Hit@2 (0.8986), Hit@5 (0.9459), and average Hit (0.8806), indicating that jointly leveraging variable and pipeline  context is most effective for top-$k$ retrieval.
The pipeline-based strategy (\kgmetaspdepip) attains the highest NDCG@1 (0.8811) and the second-best average NDCG (0.8665), while the variable-based one (\kgmetaspdevar) remains competitive, \eg the second-highest Hit@5 (0.9324).
Among non-KGE baselines, SF-Dat. is strongest leveraging pre-trained language models, particularly on NDCG metrics, indicating textual metadata from OpenML provide useful information for \dpse.

\begin{table}[t]
\centering
\caption{
    Hit@k of similarity measures according to performance-based ground truth.
    A hit is recorded when a dataset retrieved within the top-k has a ground-truth similarity with the query dataset greater than a similarity threshold ($ST$). 
    Results are reported for $ST = 0.8$ and $ST = 0.9$. 
    The highest score per metric is in bold and the second highest is underlined.
}
\vspace{-2ex}
\label{tab:results-hit}
\setlength{\tabcolsep}{1.5pt} 
    \begin{tabular}{l rrrr rrrr}
    \toprule
    & \multicolumn{4}{c}{\textbf{Hit@k} ($ST$=0.8)} & \multicolumn{4}{c}{\textbf{Hit@k} ($ST$=0.9)} \\
    \cmidrule(lr){2-5} \cmidrule(lr){6-9}
    \textbf{Method} & \textbf{k=1} & \textbf{k=2} & \textbf{k=5} & \textbf{Avg.} & \textbf{k=1} & \textbf{k=2} & \textbf{k=5} & \textbf{Avg.} \\
    \midrule
    GED~\cite{redyuk2024assisted} &
    0.5743 & 0.8243 & 0.9054 & 0.7680 &
    0.4527 & 0.7432 & 0.8514 & 0.6824 \\
    MF~\cite{feurer2015initializing} &
    0.6892 & 0.8243 & 0.8851 & 0.7996 &
    0.6622 & 0.8041 & 0.8716 & 0.7793 \\
    MF Norm.~\cite{feurer2015initializing} &
    0.6892 & \underline{0.8851} & 0.9324 & 0.8356 &
    \underline{0.7905} & 0.8446 & 0.9122 & 0.8491 \\
    SiFi-Dat.~\cite{cafarella2008webtables,zhang2018ad} &
    \underline{0.8514} & \underline{0.8851} & 0.9054 & \underline{0.8806} &
    \textbf{0.7973} & \underline{0.8649} & 0.8919 & 0.8514 \\
    SiFi-Pip.~\cite{cafarella2008webtables,zhang2018ad} &
    0.6351 & 0.8176 & \underline{0.9527} & 0.8018 &
    0.6216 & 0.7568 & \underline{0.9324} & 0.7703 \\
    MuFi-Dat.~\cite{pimplikar2012answering,zhang2018ad} &
    0.7905 & \underline{0.8851} & 0.8987 & 0.8581 &
    0.7635 & 0.8378 & 0.8784 & 0.8266 \\
    \textbf{\kgmetaspdevar} &
    0.7973 & 0.8716 & 0.9324 & 0.8671 & 
    0.7365 & 0.8581 & \underline{0.9324} & 0.8423 \\
    \textbf{\kgmetaspdepip} &
    \textbf{0.8581} & \textbf{0.9122} & 0.9460 & 0.8671 & 
    \underline{0.7905} & \underline{0.8649} & 0.9257 & \underline{0.8604} \\
    \textbf{\kgmetaspdecomb} &
    0.8176 & \textbf{0.9122} & \textbf{0.9662} & \textbf{0.8987} &
    \textbf{0.7973} & \textbf{0.8986} & \textbf{0.9459} & \textbf{0.8806} \\
    \bottomrule
    \end{tabular}
\end{table}

\begin{table}[t]
\centering
\caption{
    NDCG@k of similarity measures according to performance-based ground truth.
    The highest score per metric is in bold and the second highest is underlined.
}
\vspace{-2ex}
\label{tab:results-ndcg}
\setlength{\tabcolsep}{1.5pt} 
    \begin{tabular}{l rrrr}
    \toprule
    & \multicolumn{4}{c}{\textbf{NDCG@k}} \\
    \cmidrule(lr){2-5}
    \textbf{Method} & \textbf{k=1} & \textbf{k=2} & \textbf{k=5} & \textbf{Avg.} \\
    \midrule
    GED~\cite{redyuk2024assisted} &
    0.5684 & 0.6308 & 0.6847 & 0.6280 \\
    MF~\cite{feurer2015initializing} &
    0.6939 & 0.6874 & 0.7016 & 0.6943 \\
    MF Norm.~\cite{feurer2015initializing} &
    0.8561 & 0.8418 & 0.8358 & 0.8446 \\
    SiFi-Dat.~\cite{cafarella2008webtables,zhang2018ad} &
    \underline{0.8753} & \textbf{0.8768} & \textbf{0.8693} & \textbf{0.8738} \\
    SiFi-Pip.~\cite{cafarella2008webtables,zhang2018ad} &
    0.8255 & 0.8073 & 0.8248 & 0.8192 \\
    MuFi-Dat.~\cite{pimplikar2012answering,zhang2018ad} &
    0.8229 & 0.8338 & 0.8398 & 0.8322 \\
    \textbf{\kgmetaspdevar} &
    0.8010 & 0.8176 & 0.8136 & 0.8107 \\
    \textbf{\kgmetaspdepip} &
    \textbf{0.8811} & \underline{0.8642} & \underline{0.8541} & \underline{0.8665} \\
    \textbf{\kgmetaspdecomb} &
    0.8436 & 0.8562 & 0.8388 & 0.8462 \\
    \bottomrule
    \end{tabular}
\end{table}

\medskip
\noindent \textbf{Ablation Study: \mlseakg Integration Impact.}
To measure the effect of semantic enrichment through \mlseakg, we conducted an ablation study comparing models trained with and without \mlseakg integration across \ppe and \dpse. 
For \ppe, in the \textbf{unseen dataset} scenario (Table~\ref{tab:mlseakg-ablation-dataset}), meta-classification performance shows mixed effects: for accuracy prediction, \mlseakg increases accuracy (0.7413 vs.\ 0.7302) and F1 (0.7427 vs.\ 0.7320), whereas for precision prediction, the non-\mlseakg variant performs slightly better. 
Meta-regression results consistently favor \mlseakg. 
In the \textbf{unseen pipeline configuration} scenario (Table~\ref{tab:mlseakg-ablation-pipeline}), \mlseakg yields uniform improvements across all meta-classification evaluations, with gains of up to 0.0035 in accuracy and 0.0036 in F1. 
Meta-regression results show similar patterns: \mlseakg reduces MSE by 0.0004 for both meta-targets, and increases R\textsuperscript{2} by 0.0115 and 0.0101 respectively.
For \dpse, we selected the best-performing configuration for each condition based on NDCG@5. As shown in Table~\ref{tab:mlseakg-ablation-similarity}, integrating \mlseakg improves Hit@1 from 0.7432 to 0.7905 and NDCG@1 from 0.8551 to 0.8811, indicating stronger retrieval of the most similar datasets. Without \mlseakg, performance is slightly higher for Hit@5 and NDCG@5, suggesting that \mlseakg primarily strengthens top-rank similarity retrieval.
Overall, \mlseakg-enhanced models outperform their non-enhanced counterparts in 14 out of 16 comparisons for \ppe and in top rank comparisons for \dpse .
\mlseakg provides clear benefits for \dpse at top ranks and consistent improvements for \ppe in the unseen pipeline scenario, while offering moderate gains in unseen dataset meta-regression tasks.

\begin{table}[t]
\centering
\caption{Ablation of \mlseakg integration for \ppe for the unseen datasets. Experimental setup follows Table~\ref{tab:run-perf-pred-combined-unseen-datasets-compact}. The best score per column is in bold.}
\vspace{-2ex}
\label{tab:mlseakg-ablation-dataset}
\setlength{\tabcolsep}{3pt}
\begin{tabular}{l l rrrr}
\toprule
\textbf{Method} & \textbf{\mlseakg} &
\multicolumn{2}{c}{\textbf{Meta-Regression}} &
\multicolumn{2}{c}{\textbf{Meta-Classification}} \\
\cmidrule(lr){3-4} \cmidrule(lr){5-6}
& & \textbf{MSE} & \textbf{R2} & \textbf{Accuracy} & \textbf{F1} \\
\midrule
\multicolumn{6}{l}{\textbf{Target: Accuracy}} \\
MF All + \kgmetasppe     & With & \textbf{0.0101} & \textbf{0.6181} & \textbf{0.7413} & \textbf{0.7427} \\
MF All + \kgmetasppe & Without      & 0.0103          & 0.6114          & 0.7302          & 0.7320 \\
\midrule
\multicolumn{6}{l}{\textbf{Target: Precision}} \\
MF All + \kgmetasppe    & With & \textbf{0.0164} & \textbf{0.5449} & 0.7492          & 0.7516 \\
MF All + \kgmetasppe & Without      & 0.0167          & 0.5358          & \textbf{0.7537} & \textbf{0.7562} \\
\bottomrule
\end{tabular}
\end{table}

\begin{table}[t]
\centering
\caption{Ablation of \mlseakg integration for \ppe for the unseen pipeline configurations. Experimental setup follows Table~\ref{tab:run-perf-pred-combined-unseen-runs-compact}. The best score per column is in bold.}
\vspace{-2ex}
\label{tab:mlseakg-ablation-pipeline}
\setlength{\tabcolsep}{3pt}
\begin{tabular}{l l rrrr}
\toprule
\textbf{Method} & \textbf{\mlseakg} &
\multicolumn{2}{c}{\textbf{Meta-Regression}} &
\multicolumn{2}{c}{\textbf{Meta-Classification}} \\
\cmidrule(lr){3-4} \cmidrule(lr){5-6}
& & \textbf{MSE} & \textbf{R2} & \textbf{Accuracy} & \textbf{F1} \\
\midrule
\multicolumn{6}{l}{\textbf{Target: Accuracy}} \\
MF All + \kgmetasppe & With    & \textbf{0.0070} & \textbf{0.7361} & \textbf{0.8250} & \textbf{0.8244} \\
MF All + \kgmetasppe & Without & 0.0074          & 0.7246          & 0.8225          & 0.8220          \\
\midrule
\multicolumn{6}{l}{\textbf{Target: Precision}} \\
MF All + \kgmetasppe & With    & \textbf{0.0131} & \textbf{0.6264} & \textbf{0.8251} & \textbf{0.8247} \\
MF All + \kgmetasppe & Without & 0.0135          & 0.6163          & 0.8216          & 0.8211          \\
\bottomrule
\end{tabular}
\vspace{-2ex}
\end{table}

\begin{table}[t]
\centering
\caption{Ablation of \mlseakg integration on \dpse. For Hit@k, datasets with similarity > 0.9 are considered hits.}
\vspace{-2ex}
\label{tab:mlseakg-ablation-similarity}
\setlength{\tabcolsep}{3pt}
\begin{tabular}{l l rrr rrr}
\toprule
& & \multicolumn{3}{c}{\textbf{Hit@k}} & \multicolumn{3}{c}{\textbf{NDCG@k}} \\
\cmidrule(lr){3-5} \cmidrule(lr){6-8}
\textbf{Method} & \textbf{\mlseakg} & \textbf{k=1} & \textbf{k=2} & \textbf{k=5} & \textbf{k=1} & \textbf{k=2} & \textbf{k=5} \\
\midrule
\kgmetaspdepip & With    & \textbf{0.7905} & \textbf{0.8649} & 0.9257 & \textbf{0.8811} & \textbf{0.8642} & 0.8541 \\
\kgmetaspdepip & Without & 0.7432          & \textbf{0.8649} & \textbf{0.9527} & 0.8551 & 0.8606 & \textbf{0.8560} \\
\bottomrule
\end{tabular}
\end{table}

\section{Conclusion}
\label{sec:conclusion}

In this paper, we presented a novel approach that leverages KGEs for \ppe and \dpse. 
Our approach demonstrates robust generalization of \ppe with a single meta-model, both for unseen datasets and pipeline configurations and leading performance in \dpse. 
These results highlight the importance of incorporating pipeline configuration metadata and open experiment data as structured KG context.
Our approach is model-agnostic and applies to different dataset modalities, pipelines, and meta-learning objectives, opening new directions grounded in structured ML knowledge.

While our approach is a first demonstration of the effect of KGEs in meta-learning, its  implementation, \newframework, can be furter improved. 
First, we currently focused only on scikit-learn pipelines from OpenML to maintain a consistent representation of pipeline components.
Still, \newframework can be extended for other ML libraries or platforms, by mapping their pipeline information to the ExeKG ontology. 
Second, \newframework embeddings are static, meaning that incorporating new data requires retrofitting or re-embedding of the KG. Notably, RDF2Vec can support incremental updates~\cite{hahn2025rdf2vec}.
In future work, we plan to extend our approach to diverse ML frameworks beyond scikit-learn and investigate methods for dynamic embedding updates in evolving KGs. We also plan to test our approach for end-to-end hyper-parameter optimization.
\paragraph*{Supplemental Material Statement:}
Source code for \newframework, \newkg generation, \newbenchmark, and evaluations is available at: \url{https://github.com/dtai-kg/KGmetaSP}.

\paragraph*{Use of Generative AI:}
The authors used generative AI tools (Google Gemini and OpenAI ChatGPT) to help prepare this manuscript. These tools were employed exclusively for language refinement, restructuring of text and figures, and generation of preliminary phrasing suggestions. All conceptual contributions, methodological designs, experiments, analyses, figures, tables, and conclusions were developed solely by the authors. The authors reviewed, edited, and verified all AI-assisted text and take full responsibility for the final paper content.

\begin{credits}
\subsubsection{\ackname} The work was partially supported by the EU project SMARTY (GA 101140087), as well as Flanders Make (REXPEK project), the strategic research centre for the manufacturing industry and the Flemish Government under the “Onderzoeksprogramma Artificiële Intelligentie (AI) Vlaanderen” program.
\end{credits}

\bibliographystyle{splncs04}
\bibliography{references}

@Article{hospedales2022meta,
    author = {Hospedales, Timothy and Antoniou, Antreas and Micaelli, Paul and Storkey, Amos},
    journal = {IEEE Transactions on Pattern Analysis \& Machine Intelligence},
    title = {{Meta-Learning in Neural Networks: A Survey}},
    year = {2022},
    volume = {44},
    number = {09},
    ISSN = {1939-3539},
    pages = {5149-5169},
    keywords = {Task analysis;Optimization;Training;Machine learning algorithms;Predictive models;Neural networks;Deep learning},
    doi = {10.1109/TPAMI.2021.3079209},
    _url = {https://doi.ieeecomputersociety.org/10.1109/TPAMI.2021.3079209},
    publisher = {IEEE Computer Society},
    address = {Los Alamitos, CA, USA},
    month = sep
}

@Book{brazdil2008meta,
    author = {Brazdil, Pavel and Giraud-Carrier, Christophe and Soares, Carlos and Vilalta, Ricardo},
    title = {Metalearning: Applications to Data Mining},
    year = {2008},
    isbn = {3540732624},
    publisher = {Springer Publishing Company, Incorporated},
    edition = {1},
}

@misc{publioMLSchemaExposingSemantics2018,
  title = {{{ML-Schema}}: {{Exposing}} the {{Semantics}} of {{Machine Learning}} with {{Schemas}} and {{Ontologies}}},
  shorttitle = {{{ML-Schema}}},
  author = {Publio, Gustavo Correa and Esteves, Diego and {\L}awrynowicz, Agnieszka and Panov, Pan{\v c}e and Soldatova, Larisa and Soru, Tommaso and Vanschoren, Joaquin and Zafar, Hamid},
  year = {2018},
  month = jul,
  number = {arXiv:1807.05351},
  eprint = {1807.05351},
  primaryclass = {cs, stat},
  publisher = {arXiv},
  urldate = {2024-08-28},
  archiveprefix = {arXiv}
}

@inproceedings{NIPS2013_1cecc7a7,
  title = {Translating Embeddings for Modeling Multi-Relational Data},
  booktitle = {Advances in Neural Information Processing Systems},
  author = {Bordes, Antoine and Usunier, Nicolas and {Garcia-Duran}, Alberto and Weston, Jason and Yakhnenko, Oksana},
  year = {2013},
  volume = {26},
  publisher = {{Curran Associates, Inc.}}
}

@article{zheng2021pharmkg,
  title= {PharmKG: a dedicated knowledge graph benchmark for bomedical data mining},
  author= {Zheng, Shuangjia and Rao, Jiahua and Song, Ying and Zhang, Jixian and Xiao, Xianglu and Fang, Evandro Fei and Yang, Yuedong and Niu, Zhangming},
  journal= {Briefings in bioinformatics},
  volume= {22},
  number= {4},
  pages= {bbaa344},
  year= {2021},
  publisher= {Oxford University Press}
}

@article{ristoski2019rdf2vec,
  title= {RDF2Vec: RDF graph embeddings and their applications},
  author= {Ristoski, Petar and Rosati, Jessica and Di Noia, Tommaso and De Leone, Renato and Paulheim, Heiko},
  journal= {Semantic Web},
  volume= {10},
  number= {4},
  pages= {721--752},
  year= {2019},
  publisher= {IOS Press}
}

@inproceedings{ristoski2016rdf2vec,
  title= {Rdf2vec: Rdf graph embeddings for data mining},
  author= {Ristoski, Petar and Paulheim, Heiko},
  booktitle= {International semantic web conference},
  pages= {498--514},
  year= {2016},
  organization= {Springer}
}

@article{wu2020comprehensive,
  title= {A comprehensive survey on graph neural networks},
  author= {Wu, Zonghan and Pan, Shirui and Chen, Fengwen and Long, Guodong and Zhang, Chengqi and Philip, S Yu},
  journal= {IEEE Trans. Neural Netw.},
  volume= {32},
  number= {1},
  pages= {4--24},
  year= {2020},
  publisher= {IEEE}
}

@article{OpenML2013,
author = {Vanschoren, Joaquin and van Rijn, Jan N. and Bischl, Bernd and Torgo, Luis},
title = {OpenML: Networked Science in Machine Learning},
journal = {SIGKDD Expl.},
volume = {15},
number = {2},
year = {2013},
pages = {49--60},
publisher = {ACM},
address = {New York, NY, USA},
}

@article{redyuk2024assisted,
  title= {Assisted design of data science pipelines},
  author= {Redyuk, Sergey and Kaoudi, Zoi and Schelter, Sebastian and Markl, Volker},
  journal= {The VLDB Journal},
  pages= {1--25},
  year= {2024},
  publisher= {Springer}
}

@inproceedings{klironomos2023exekglib,
  title= {ExeKGLib: knowledge graphs-empowered machine learning analytics},
  author= {Klironomos, Antonis and Zhou, Baifan and Tan, Zhipeng and Zheng, Zhuoxun and Mohamed, Gad-Elrab and Paulheim, Heiko and Kharlamov, Evgeny},
  booktitle= {ESWC},
  pages= {123--127},
  year= {2023},
  organization= {Springer}
}

@InProceedings{dasoulas2024mlsea,
author="Dasoulas, Ioannis
and Yang, Duo
and Dimou, Anastasia",
title="MLSea: A Semantic Layer for Discoverable Machine Learning",
booktitle="The Semantic Web",
year="2024",
_publisher="Springer Nature Switzerland",
publisher="CSAL",
address="Cham",
pages="178--198",
abstract="With the Machine Learning (ML) field rapidly evolving, ML pipelines continuously grow in numbers, complexity and components. Online platforms (e.g., OpenML, Kaggle) aim to gather and disseminate ML experiments. However, available knowledge is fragmented with each platform representing distinct components of the ML process or intersecting components but in different ways. To address this problem, we leverage semantic web technologies to model and integrate ML datasets, experiments, software and scientific works into MLSea, a resource consisting of: (i)MLSO, an ontology that models ML datasets, pipelines and implementations; (ii)MLST, taxonomies with collections of ML knowledge formulated as controlled vocabularies; and (iii) MLSea-KG, an RDF graph containing ML datasets, pipelines, implementations and scientific works from diverse sources. MLSea paves the way for improving the search, explainability and reproducibility of ML pipelines.",
isbn="978-3-031-60635-9"
}

@TechReport{dcat,
  author      = {Maali, Fadi and Erickson, John},
  institution = {World Wide Web Consortium (W3C)},
  title       = {{Data Catalog Vocabulary (DCAT)}},
  year        = {2014},
  month       = jan,
  type        = {Recommendation},
}

@inproceedings{zheng2022executable,
  title= {Executable knowledge graphs for machine learning: a Bosch case of welding monitoring},
  author= {Zheng, Zhuoxun and Zhou, Baifan and Zhou, Dongzhuoran and Zheng, Xianda and Cheng, Gong and Soylu, Ahmet and Kharlamov, Evgeny},
  booktitle= {International Semantic Web Conference},
  pages= {791--809},
  year= {2022},
  organization= {Springer}
}

@InProceedings{leite2021exploiting,
  title = 	 {{Exploiting Performance-based Similarity between Datasets in Metalearning}},
  author =       {Leite, Rui and Brazdil, Pavel},
  booktitle = 	 {MetaLearning@AAAI2021},
  pages = 	 {90--99},
  year = 	 {2021},
  _editor = 	 {Guyon, Isabelle and van Rijn, Jan N. and Treguer, Sébastien and Vanschoren, Joaquin},
  _volume = 	 {140},
  series = 	 {PMLR},
  _month = 	 {09 Feb},
  publisher =    {PMLR},
  pdf = 	 {http://proceedings.mlr.press/v140/leite21a/leite21a.pdf},
  _url = 	 {https://proceedings.mlr.press/v140/leite21a.html}
}

@article{preisner2023universal,
  title= {Universal preprocessing operators for embedding knowledge graphs with literals},
  author= {Preisner, Patryk and Paulheim, Heiko},
  journal= {arXiv preprint arXiv:2309.03023},
  year= {2023}
}

@inproceedings{zhang2018ad, author = {Zhang, Shuo and Balog, Krisztian}, title = {{Ad Hoc Table Retrieval using Semantic Similarity}}, year = {2018}, isbn = {9781450356398}, publisher = {International World Wide Web Conferences Steering Committee}, address = {Republic and Canton of Geneva, CHE}, _url = {https://doi.org/10.1145/3178876.3186067},  booktitle = {WWW18}, pages = {1553–1562}, numpages = {10}, keywords = {semantic matching, semantic representations, semantic similarity, table retrieval, table search}, location = {Lyon, France}, series = {WWW '18} }

@article{stolte2024methods,
   title= {{Methods for quantifying dataset similarity: a review, taxonomy and comparison}},
   volume= {18},
   ISSN= {1935-7516},
   _url= {http://dx.doi.org/10.1214/24-SS149},
   _DOI= {10.1214/24-ss149},
   number= {none},
   journal= {Statistics Surveys},
   publisher= {Institute of Mathematical Statistics},
   author= {Stolte, Marieke and Kappenberg, Franziska and Rahnenführer, Jörg and Bommert, Andrea},
   year= {2024},
   month=jan }

@Article{feurer2015initializing, 
    title = {Initializing Bayesian Hyperparameter Optimization via Meta-Learning}, volume = {29}, 
    url = {https://ojs.aaai.org/index.php/AAAI/article/view/9354}, 
    DOI = {10.1609/aaai.v29i1.9354}, 
    number = {1}, 
    journal = {Proceedings of the AAAI Conference on Artificial Intelligence}, 
    author = {Feurer, Matthias and Springenberg, Jost and Hutter, Frank}, 
    year = {2015}, 
    month = {Feb.}
}

@inproceedings{bensusan2000discovering,
  title= {{Discovering Task Neighbourhoods Through Landmark Learning Performances}},
  author= {Hilan Bensusan and Christophe G. Giraud-Carrier},
  booktitle= {ECML-PKDD},
  year= {2000},
  _url= {https://api.semanticscholar.org/CorpusID:749402},
  _doi = {10.1007/3-540-45372-5_32}
}

@article{jooma2021dataset,
  author       = {Hadi S. Jomaa and
                  Lars Schmidt{-}Thieme and
                  Josif Grabocka},
  title        = {{Dataset2Vec: learning dataset meta-features}},
  journal      = {Data Min. Knowl. Discov.},
  volume       = {35},
  number       = {3},
  pages        = {964--985},
  year         = {2021},
  _url          = {https://doi.org/10.1007/s10618-021-00737-9},
  _doi          = {10.1007/S10618-021-00737-9},
  timestamp    = {Tue, 07 May 2024 20:27:49 +0200},
  biburl       = {https://dblp.org/rec/journals/datamine/JomaaSG21.bib},
  bibsource    = {dblp computer science bibliography, https://dblp.org}
}

@misc{kim2018learning,
      title= {{Learning to Warm-Start Bayesian Hyperparameter Optimization}}, 
      author= {Jungtaek Kim and Saehoon Kim and Seungjin Choi},
      year= {2018},
      eprint= {1710.06219},
      archivePrefix= {arXiv},
      primaryClass= {stat.ML},
      _url= {https://arxiv.org/abs/1710.06219}, 
}

@article{pedregosa2011scikit,
  title= {{Scikit-learn: Machine learning in Python}},
  author= {Pedregosa, Fabian and Varoquaux, Ga{\"e}l and Gramfort, Alexandre and Michel, Vincent and Thirion, Bertrand and Grisel, Olivier and Blondel, Mathieu and Prettenhofer, Peter and Weiss, Ron and Dubourg, Vincent and others},
  journal= {the Journal of machine Learning research},
  volume= {12},
  pages= {2825--2830},
  year= {2011},
  publisher= {JMLR. org}
}

@article{cafarella2008webtables, author = {{Cafarella, Michael J. and Halevy, Alon and Wang, Daisy Zhe and Wu, Eugene and Zhang, Yang}}, title = {WebTables: exploring the power of tables on the web}, year = {2008}, issue_date = {August 2008}, publisher = {VLDB Endowment}, volume = {1}, number = {1}, issn = {2150-8097}, _url = {https://doi.org/10.14778/1453856.1453916}, _doi = {10.14778/1453856.1453916}, journal = {Proc. VLDB Endow.}, month = aug, pages = {538–549}, numpages = {12} }

@inproceedings{reimers2019sentence,
    title = "Sentence-{BERT}: Sentence Embeddings using {S}iamese {BERT}-Networks",
    author = "Reimers, Nils  and
      Gurevych, Iryna",
    _editor = "Inui, Kentaro  and
      Jiang, Jing  and
      Ng, Vincent  and
      Wan, Xiaojun",
    booktitle = "EMNLP-IJCNLP 2019",
    month = nov,
    year = "2019",
    address = "Hong Kong, China",
    publisher = "Association for Computational Linguistics",
    _url = "https://aclanthology.org/D19-1410/",
    _doi = "10.18653/v1/D19-1410",
    pages = "3982--3992"
}

@article{pimplikar2012answering, author = {Pimplikar, Rakesh and Sarawagi, Sunita}, title = {Answering table queries on the web using column keywords}, year = {2012}, issue_date = {June 2012}, publisher = {VLDB Endowment}, volume = {5}, number = {10}, issn = {2150-8097}, _url = {https://doi.org/10.14778/2336664.2336665}, _doi = {10.14778/2336664.2336665}, journal = {Proc. VLDB Endow.}, month = jun, pages = {908–919}, numpages = {12} }

@inproceedings{ogilvie2003combining, author = {Ogilvie, Paul and Callan, Jamie}, title = {Combining document representations for known-item search}, year = {2003}, isbn = {1581136463}, publisher = {Association for Computing Machinery}, address = {New York, NY, USA}, _url = {https://doi.org/10.1145/860435.860463}, _doi = {10.1145/860435.860463}, booktitle = {SIGIR03}, pages = {143–150}, numpages = {8}, keywords = {meta-search algorithms, language models, known-item finding, data fusion}, location = {Toronto, Canada}, series = {SIGIR '03} }

@inproceedings{coates-bollegala-2018-frustratingly,
    title = "Frustratingly Easy Meta-Embedding {--} Computing Meta-Embeddings by Averaging Source Word Embeddings",
    author = "Coates, Joshua  and
      Bollegala, Danushka",
    editor = "Walker, Marilyn  and
      Ji, Heng  and
      Stent, Amanda",
    booktitle = "Proceedings of the 2018 Conference of the North {A}merican Chapter of the Association for Computational Linguistics: Human Language Technologies, Volume 2 (Short Papers)",
    month = jun,
    year = "2018",
    address = "New Orleans, Louisiana",
    publisher = "Association for Computational Linguistics",
    url = "https://aclanthology.org/N18-2031/",
    doi = "10.18653/v1/N18-2031",
    pages = "194--198"
}

@article{bilalli2017on,
author = {Bilalli, Besim and Abelló, Alberto and Aluja-Banet, Tomàs},
doi = {10.1515/amcs-2017-0048},
url = {https://doi.org/10.1515/amcs-2017-0048},
title = {{On the predictive power of meta-features in OpenML}},
journal = {International Journal of Applied Mathematics and Computer Science},
number = {4},
volume = {27},
year = {2017},
pages = {697--712}
}

@inproceedings{schlichtkrull2018modeling,
  title= {Modeling relational data with graph convolutional networks},
  author= {Schlichtkrull, Michael and Kipf, Thomas N and Bloem, Peter and van den Berg, Rianne and Titov, Ivan and Welling, Max},
  booktitle= {European semantic web conference},
  pages= {593--607},
  year= {2018},
  organization= {Springer}
}

@book{10.5555/212782,
editor = {Michie, Donald and Spiegelhalter, D. J. and Taylor, C. C. and Campbell, John},
title = {Machine learning, neural and statistical classification},
year = {1995},
isbn = {013106360X},
publisher = {Ellis Horwood},
address = {USA}
}

@article{nickel2016review,
  author= {Nickel, Maximilian and Murphy, Kevin and Tresp, Volker and Gabrilovich, Evgeniy},
  journal= {Proceedings of the IEEE}, 
  title= {A Review of Relational Machine Learning for Knowledge Graphs}, 
  year= {2016},
  volume= {104},
  number= {1},
  pages= {11-33},
  doi= {10.1109/JPROC.2015.2483592}
}

@inproceedings{alcobaca2025exploring,
  author    = {Alcobaça, Edesio and De Carvalho, André C. P. L. F.},
  title     = {Exploring One Million Machine Learning Pipelines: A Benchmarking Study},
  booktitle = {AutoML 2025},
  year      = {2025}
}

@article{olson2017pmlb,
  author  = {Olson, R. S. and La Cava, W. and Orzechowski, P. and Urbanowicz, R. J. and Moore, J. H.},
  title   = {PMLB: a large benchmark suite for machine learning evaluation and comparison},
  journal = {BioData mining},
  volume  = {10},
  pages   = {1--13},
  year    = {2017}
}

@article{gijsbers2024amlb,
  author  = {Gijsbers, P. and others},
  title   = {Amlb: an automl benchmark},
  journal = {Journal of Machine Learning Research},
  volume  = {25},
  pages   = {1--65},
  year    = {2024}
}

@article{carneiro2021using,
  title= {Using meta-learning to predict performance metrics in machine learning problems},
  author= {Carneiro, Davide and Guimar{\~a}es, Miguel and Carvalho, Mariana and Novais, Paulo},
  journal= {Expert Systems},
  year= {2021},
  publisher= {Wiley Online Library},
  doi={https://doi.org/10.1111/exsy.12900}
}

@inproceedings{feurer2014using,
author = {Feurer, Matthias and Springenberg, Jost Tobias and Hutter, Frank},
title = {{Using Meta-learning to Initialize Bayesian Optimization of Hyperparameters}},
year = {2014},
isbn = {16130073},
publisher = {CEUR-WS.org},
address = {Aachen, DEU},
booktitle = {Proceedings of the 2014 International Conference on Meta-Learning and Algorithm Selection - Volume 1201},
pages = {3–10},
numpages = {8},
location = {Prague, Czech Republic},
series = {MLAS'14}
}

@InProceedings{castiello2005metadata,
author={Castiello, Ciro
and Castellano, Giovanna
and Fanelli, Anna Maria},
editor={Torra, Vicen{\c{c}}
and Narukawa, Yasuo
and Miyamoto, Sadaaki},
title={{Meta-data: Characterization of Input Features for Meta-learning}},
booktitle={Modeling Decisions for Artificial Intelligence},
year={2005},
publisher={Springer Berlin Heidelberg},
address={Berlin, Heidelberg},
pages={457--468},
isbn={978-3-540-31883-5},
doi={https://doi.org/10.1007/11526018_45}
}

@InProceedings{kalousis2001feature,
author="Kalousis, Alexandros
and Hilario, Melanie",
editor="Cheung, David
and Williams, Graham J.
and Li, Qing",
title="Feature Selection for Meta-learning",
booktitle="Advances in Knowledge Discovery and Data Mining",
year="2001",
publisher="Springer Berlin Heidelberg",
address="Berlin, Heidelberg",
pages="222--233",
isbn="978-3-540-45357-4",
doi="https://doi.org/10.1007/3-540-45357-1_26"
}

@article{lemke2013metalearning,
author = {Lemke, Christiane and Budka, Marcin and Gabrys, Bogdan},
year = {2013},
month = {06},
pages = {},
title = {{Metalearning: a survey of trends and technologies}},
volume = {DOI: 10.1007/s10462-013-9406-y},
journal = {Artificial Intelligence Review},
doi = {10.1007/s10462-013-9406-y}
}

@InProceedings{palumbo2023real,
author="Palumbo, Guilherme
and Guimar{\~a}es, Miguel
and Carneiro, Davide
and Novais, Paulo
and Alves, Victor",
editor="Juli{\'a}n, Vicente
and Carneiro, Jo{\~a}o
and Alonso, Ricardo S.
and Chamoso, Pablo
and Novais, Paulo",
title="Real-Time Algorithm Recommendation Using Meta-Learning",
booktitle="Ambient Intelligence---Software and Applications---13th International Symposium on Ambient Intelligence",
year="2023",
publisher="Springer International Publishing",
address="Cham",
pages="249--258",
isbn="978-3-031-22356-3",
doi="10.1007/978-3-031-22356-3_24"
}

@inproceedings{drori2021alpha3dm,
	AUTHOR = {Drori, Iddo and Krishnamurthy, Yamuna and Rampin, Remi and DE PAULA LOURENCO, Raoni and Piazentin Ono, Jorge and Cho, Kyunghyun and Silva, Claudio and Freire, Juliana},
	EPRINT = {https://orbilu.uni.lu/10993/57634},
	EPRINTTYPE = {hdl},
	TITLE = {AlphaD3M: Machine Learning Pipeline Synthesis},
	LANGUAGE = {English},
	YEAR = {2021},
	Booktitle = {ICML 2018 AutoML Workshop}
}

@inproceedings{pineda2023deeppipe,
author = {Pineda Arango, Sebastian and Grabocka, Josif},
title = {Deep Pipeline Embeddings for AutoML},
year = {2023},
isbn = {9798400701030},
publisher = {Association for Computing Machinery},
address = {New York, NY, USA},
url = {https://doi.org/10.1145/3580305.3599303},
doi = {10.1145/3580305.3599303},
booktitle = {Proceedings of the 29th ACM SIGKDD Conference on Knowledge Discovery and Data Mining},
pages = {1907–1919},
numpages = {13},
location = {Long Beach, CA, USA},
series = {KDD '23}
}

@inproceedings{chen2025benchmarking,
author = {Chen, Qiaosheng and Huang, Kaijia and Zhou, Xiao and Luo, Weiqing and Cui, Yuanning and Cheng, Gong},
title = {Benchmarking Recommendation, Classification, and Tracing Based on Hugging Face Knowledge Graph},
year = {2025},
isbn = {9798400715921},
publisher = {Association for Computing Machinery},
address = {New York, NY, USA},
url = {https://doi.org/10.1145/3726302.3730277},
doi = {10.1145/3726302.3730277},
booktitle = {Proceedings of the 48th International ACM SIGIR Conference on Research and Development in Information Retrieval},
pages = {3433–3443},
numpages = {11},
location = {Padua, Italy},
series = {SIGIR '25}
}

@article{liu2023taskoriented,
author = {Liu, Mingwei and Zhao, Chengyuan and Peng, Xin and Yu, Simin and Wang, Haofen and Sha, Chaofeng},
title = {Task-Oriented ML/DL Library Recommendation Based on a Knowledge Graph},
year = {2023},
issue_date = {Aug. 2023},
publisher = {IEEE Press},
volume = {49},
number = {8},
issn = {0098-5589},
url = {https://doi.org/10.1109/TSE.2023.3285280},
doi = {10.1109/TSE.2023.3285280},
journal = {IEEE Trans. Softw. Eng.},
month = aug,
pages = {4081–4096},
numpages = {16}
}

@inproceedings{hahn2025rdf2vec,
author = {Hahn, Sang Hyu and Paulheim, Heiko},
title = {RDF2vec Embeddings for Updateable Knowledge Graphs – Reuse, Don’t Retrain!},
year = {2025},
isbn = {978-3-031-78951-9},
publisher = {Springer-Verlag},
address = {Berlin, Heidelberg},
url = {https://doi.org/10.1007/978-3-031-78952-6_30},
doi = {10.1007/978-3-031-78952-6_30},
booktitle = {The Semantic Web: ESWC 2024 Satellite Events: Hersonissos, Crete, Greece, May 26–30, 2024, Proceedings, Part I},
pages = {217–222},
numpages = {6},
location = {Hersonissos, Greece}
}
\appendix
\section{Implementation Details and Additional Experiments}
\label{sec:implementation-details}


\subsection{Experimental Configuration}


\medskip
\noindent \textbf{Embedding Hyper-parameters.}
\newkg is embedded using \rdftvec with \texttt{gensim} Word2Vec and MKGA discretization.
We generate 10 walks per entity (length 20) and train Word2Vec for 10 epochs (dim=100, window=5, negative=5, min\_count=0).
Numeric literals are processed via MKGA strategy \texttt{propConst\_LOF\_del\_numbers}.
All unspecified parameters use \texttt{gensim} defaults.



\medskip
\noindent \textbf{Data Splitting for \ppe.}
We use a 70/30 train-test split (seed=42). 
Splitting is performed at the dataset level for unseen datasets or at the pipeline configuration level for unseen pipelines, such that all dataset - pipeline pairs associated with a given dataset or pipeline configuration are assigned entirely to either the training or the test split.

\medskip
\noindent \textbf{\ppe Meta-Models.}
Random Forests (scikit-learn) are used for meta-regression and meta-classification.
Hyper-parameters are selected via 10-fold CV on the training set.
Grid search space:
n\_estimators $\in$ [50-200] (clf), [20-200] (regr);
max\_depth $\in$ \{10,20,None\};
min\_samples\_split $\in$ \{2,5,10\}.
Other parameters use defaults (seed=42).

\subsection{Analysis of Computational Cost}
\label{sec:computational-cost-analysis}


\medskip
\noindent \textbf{Runtime.}
Experiments ran on an AMD EPYC-7643 CPU.
\newkg construction (2,616 \exekgs, $\sim$4.5M triples) requires $\sim$7.5h using 10 cores.
\rdftvec training requires $\sim$24h (10-core parallel walk generation).
Since a single embedding model serves all datasets and pipelines, this cost is incurred once.
\ppe training requires $\sim$9.5h.
Embedding training dominates total runtime.


\medskip
\noindent \textbf{Memory.}
KG construction peaks at $\sim$6GB RAM.
Embedding training requires $\sim$80GB RAM (walk corpus $\sim$1.5M nodes).
\ppe training requires $\sim$13GB RAM.


\medskip
\noindent \textbf{Scalability.}
\newkg construction scales linearly with the number of experiments.
\rdftvec scales with node count, walk parameters, and embedding dimension.
\ppe scales with dataset - pipeline pairs and feature dimensionality.
Embedding training is the dominant cost.

\subsection{Additional Experiments}
\label{sec:appendix-lp-ppe}


\medskip
\noindent \textbf{Experimental Setup.}
We evaluate TransE, DistMult, and ComplEx (PyKEEN) link prediction models.
Embeddings (dim=128) are trained for 1500 epochs using self-adversarial negative sampling (batch=2048, lr=0.0005, 3 negatives, margin=50).
Pipeline embeddings are aggregated and concatenated with MF All for a single pipeline-agnostic \ppe meta-model.

\medskip
\noindent \textbf{Results.}
Across both scenarios, \ie \textbf{unseen datasets} (Table~\ref{tab:appendix-perf-pred-unseen-datasets}) and \textbf{unseen pipelines} (Table~\ref{tab:appendix-perf-pred-unseen-pipelines}), \rdftvec embeddings outperform LP-based embeddings. Whenever \rdftvec surpasses the baselines, the best-performing LP model lies in between, \ie Baselines $<$ \kgmetasppe (best LP model) $<$ \kgmetasppe (\rdftvec). In cases where the strongest baselines remain superior, KG-based variants preserve the same internal ordering, \ie \kgmetasppe (best LP model) $<$ \kgmetasppe (\rdftvec).
Thus, the results reinforce two conclusions: (i) performance improvements in \ppe are primarily driven by the semantic KG representation of pipelines and datasets, and (ii) walk-based embeddings offer a structural advantage for complex ML metadata graphs such as \newkg.

\begin{table}[t]
\centering
\caption{\ppe results for unseen datasets. 
Setup follows Table~\ref{tab:run-perf-pred-combined-unseen-datasets-compact}.
}
\vspace{-3ex}
\label{tab:appendix-perf-pred-unseen-datasets}
\setlength{\tabcolsep}{2pt} 
    \begin{tabular}{@{}lllcccc@{}}
    \toprule
    \textbf{Dataset} & \textbf{Pipeline} & \textbf{Meta-} & \multicolumn{2}{c}{\textbf{Meta-Regr.}} & \multicolumn{2}{c}{\textbf{Meta-Classif.}} \\
    \cmidrule(lr){4-5} \cmidrule(lr){6-7}
    \textbf{Emb.} & \textbf{Strategy} & \textbf{Models} & \textbf{MSE} & \textbf{R2} & \textbf{Accuracy} & \textbf{F1} \\
    \midrule
    \multicolumn{6}{l}{\textbf{Target: Accuracy}} \\
    \multirow{5}{*}{MF All} & Conf.-specific & 1028 & \textbf{\underline{0.0081}} & \textbf{\underline{0.6748}} & 0.7363 & 0.7358 \\
    & \kgmetasppe (ComplEx) & 1 & 0.0106 & 0.6006 & 0.7343 & 0.7361 \\
    & \kgmetasppe (TransE) & 1 & 0.0105 & 0.6032 & 0.7351 & 0.7368 \\
    & \kgmetasppe (DistMult) & 1 & 0.0103 & 0.6104 & 0.7380 & 0.7393 \\
    & \textbf{\kgmetasppe (\rdftvec)} & 1 & 0.0101 & 0.6181 & \textbf{\underline{0.7413}} & \textbf{\underline{0.7427}} \\
    \midrule
    \multicolumn{6}{l}{\textbf{Target: Precision}} \\
    \multirow{5}{*}{MF All} & Conf.-specific & 1028 & \textbf{\underline{0.0133}} & \textbf{\underline{0.6311}} & 0.7347 & 0.7400 \\
    & \kgmetasppe (ComplEx) & 1 & 0.0175 & 0.5138 & 0.7359 & 0.7383 \\
    & \kgmetasppe (TransE) & 1 & 0.0176 & 0.5117 & 0.7430 & 0.7457 \\
    & \kgmetasppe (DistMult) & 1 & 0.0173 & 0.5198 & 0.7461 & 0.7482 \\
    & \textbf{\kgmetasppe (\rdftvec)} & 1 & 0.0164 & 0.5449 & \textbf{\underline{0.7537}} & \textbf{\underline{0.7562}} \\
    \bottomrule
    \end{tabular}
    \vspace{-1ex}
\end{table}

\begin{table}[t]
\centering
\caption{\ppe results for unseen pipeline configurations.
Setup follows Table~\ref{tab:run-perf-pred-combined-unseen-runs-compact}.
}
\vspace{-3ex}
\label{tab:appendix-perf-pred-unseen-pipelines}
\setlength{\tabcolsep}{3pt} 
    \begin{tabular}{lcccc}
    \toprule
    \textbf{Method} & \multicolumn{2}{c}{\textbf{Meta-Regr.}} & \multicolumn{2}{c}{\textbf{Meta-Classif.}} \\
    \cmidrule(lr){2-3} \cmidrule(lr){4-5}
    & \textbf{MSE} & \textbf{R2} & \textbf{Accuracy} & \textbf{F1} \\
    \midrule
    \multicolumn{5}{l}{\textbf{Target: Accuracy}} \\
    Average performance & 0.0267 & -0.0005 & 0.3303 & 0.1640 \\
    Closest embedding & 0.0127 & 0.5241 & 0.7748 & 0.7747 \\
    MF All + \kgmetasppe (ComplEx) & 0.0083 & 0.6911 & 0.8072 & 0.8062 \\
    MF All + \kgmetasppe (TransE) & 0.0081 & 0.6976 & 0.8092 & 0.8083 \\
    MF All + \kgmetasppe (DistMult) & 0.0080 & 0.7007 & 0.8076 & 0.8068 \\
    \textbf{MF All + \kgmetasppe (\rdftvec)} & \textbf{0.0070} & \textbf{0.7361} & \textbf{0.8250} & \textbf{0.8244} \\
    \midrule
    \multicolumn{5}{l}{\textbf{Target: Precision}} \\
    Average performance & 0.0352 & -0.0012 & 0.3323 & 0.1658 \\
    Closest embedding & 0.0225 & 0.3607 & 0.7773 & 0.7771 \\
    MF All + \kgmetasppe (ComplEx) & 0.0153 & 0.5650 & 0.8088 & 0.8080 \\
    MF All + \kgmetasppe (TransE) & 0.0148 & 0.5792 & 0.8072 & 0.8064 \\
    MF All + \kgmetasppe (DistMult) & 0.0148 & 0.5804 & 0.8089 & 0.8083 \\
    \textbf{MF All + \kgmetasppe (\rdftvec)} & \textbf{0.0131} & \textbf{0.6264} & \textbf{0.8251} & \textbf{0.8247} \\
    \bottomrule
    \end{tabular}
    \vspace{-6ex}
\end{table}

\end{document}